\def\year{2022}\relax

\documentclass[letterpaper]{article} 
\usepackage{aaai22}  
\usepackage{times}  
\usepackage{helvet}  
\usepackage{courier}  
\usepackage[hyphens]{url}  
\usepackage{graphicx} 
\usepackage{natbib}  
\usepackage{caption} 
\DeclareCaptionStyle{ruled}{labelfont=normalfont,labelsep=colon,strut=off} 
\usepackage{algorithm}
\usepackage{algorithmic}
\usepackage{multirow}
\usepackage{amsfonts}
\usepackage{amsmath}
\usepackage{cite}
\usepackage{booktabs}
\usepackage{subfigure}

\usepackage{newfloat}
\usepackage{listings}
\floatstyle{ruled}
\newfloat{listing}{tb}{lst}{}
\floatname{listing}{Listing}

\title{Pose-guided Feature Disentangling for Occluded Person \\ Re-identification Based on Transformer}
\author{
    Tao Wang, Hong Liu\thanks{Corresponding author}, Pinhao Song, Tianyu Guo, Wei Shi
}
\affiliations{
    Key Laboratory of Machine Perception\\
    
    Shenzhen Graduate School, Peking University\\
    \{taowang, levigty\}@stu.pku.edu.cn, \{hongliu, pinhaosong, pkusw\}@pku.edu.cn
}

\begin{document}
\maketitle

\begin{abstract}
Occluded person re-identification is a challenging task as human body parts could be occluded by some obstacles (e.g. trees, cars, and pedestrians) in certain scenes. Some existing pose-guided methods solve this problem by aligning body parts according to graph matching, but these graph-based methods are not intuitive and complicated. Therefore, we propose a transformer-based Pose-guided Feature Disentangling (PFD) method by utilizing pose information to clearly disentangle semantic components (e.g. human body or joint parts) and selectively match non-occluded parts correspondingly. First, Vision Transformer (ViT) is used to extract the patch features with its strong capability. Second, to preliminarily disentangle the pose information from patch information, the matching and distributing mechanism is leveraged in Pose-guided Feature Aggregation (PFA) module. Third, a set of learnable semantic views are introduced in transformer decoder to implicitly enhance the disentangled body part features. However, those semantic views are not guaranteed to be related to the body without additional supervision. Therefore, Pose-View Matching (PVM) module is proposed to explicitly match visible body parts and automatically separate occlusion features. Fourth, to better prevent the interference of occlusions, we design a Pose-guided Push Loss to emphasize the features of visible body parts. Extensive experiments over five challenging datasets for two tasks (occluded and holistic Re-ID) demonstrate that our proposed PFD is superior promising, which performs favorably against state-of-the-art methods. Code is available at \url{https://github.com/WangTaoAs/PFD_Net}
\end{abstract}

\begin{figure}[htb]
    \centering
    \includegraphics[height=7cm, width=7cm]{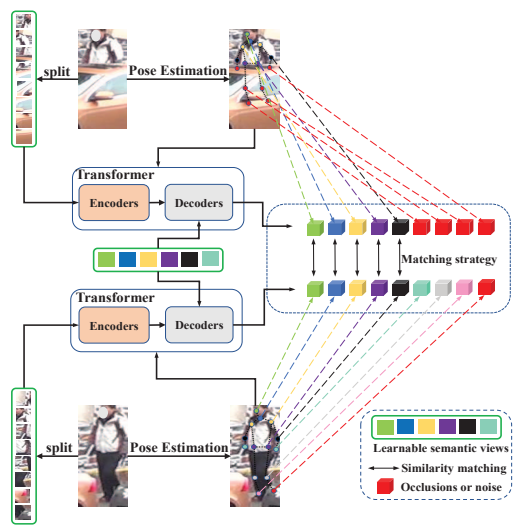}
    \caption{Illustration of Pose-guided Feature Disentangling (PFD) method in occluded person Re-ID. PFD represents an occluded person image by using the transformer to implicitly disentangle disciminative features and explicitly using pose information to guide the separation of the non-occluded features and occluded features.}
    \label{framework_easy}
\end{figure}

\section{Introduction}
Person Re-Identification (Re-ID) aims to identify a specific person across multiple non-overlapping cameras \cite{ person_reid}. It is an important subject in the field of computer vision and has a wide range of application backgrounds, such as video surveillance, activity analysis, security, and smart city. In recent years, holistic person Re-ID achieves great progress, and various of methods \cite{sun2018beyond, shi2020identity,introduction_progress_3} have been proposed. However, in real scenes, such as stations, airports, shopping malls, person can be easily occluded by some obstacles (e.g., trees, cars, pedestrians), it is difficult to match people with incomplete and invisible body parts. Therefore, the occluded person re-identification task \cite{zhuo2018occluded, miao2019pose, DRL_Net} is of important practical significance.

Compared with holistic person Re-ID, there are two major challenges for occluded person Re-ID task: (1) Due to the existence of occlusion, various noises have been introduced that result in mismatching. (2) The occlusion may have features similar to human body parts, leading to the failure of feature learning. Some early methods \cite{miao2019pose} utilize pose information to indicate non-occluded body parts on the spatial feature map and directly divide global features into partial features. These methods are intuitive but requires strict spatial feature alignment. Some recent pose-guided methods \cite{PVPM, HOReID} use graph-based approaches to model topology information by learning node-to-node or edge-to-edge correspondence to further mine the visible parts. However, these methods still suffer the problem mentioned in the challenge (2). Therefore, in this paper, to solve the above problems we explore the possibility of combining additional pose clues with transformers without spatial alignment. As the Figure \ref{framework_easy} shows, we propose PFD, a Pose-guided Feature Disentangling transformer network that utilizes pose information to clearly disentangle semantic components (e.g. human body or joint parts), and force the similarities between the occluded features and the non-occluded features to be as inconsistent as possible, which could strengthen the learning of discriminative features while reducing background noise to solve the problem of challenge (1), and effectively alleviates the failure of feature learning mentioned in challenge (2).

Specifically, the proposed PFD includes a visual context transformer encoder, a pose estimator, a Pose-guided Feature Aggregation (PFA) module, a part view based transformer decoder, and a Pose-View Matching (PVM) module. In the visual context transformer encoder, we adopt a transformer based image classification model (i.e., ViT \cite{dosovitskiy2020image}) and the camera perspective information to capture the robust global context information. PFA is developed to embed the pose information into the global context features and part features. The features obtained from PFA could preliminarily indicate visible body parts. In part view based transformer decoder, a set of learnable semantic views are introduced to implicitly enhance the disentangled body part features. Each part view feature corresponds to the discriminative part of the occlusion image. However, without additional supervision, we can only learn features implicitly and cannot constrain the learnable semantic views to capture accurate parts of the human body. Therefore, we propose a Pose-View Matching (PVM) module, which implicitly learns discriminative features and explicitly matches visible body parts, thereby separating the human body features from the occluded features and reducing the interference of noise mentioned in challenge (1). In addition, to avoid the failure of feature learning mentioned in challenge (2), we design a Pose-guided Push Loss to reduce the similarity between human body features and occlusion features.

The main contributions of this paper can be summarized as the following:
\begin{itemize}
    \item [\textbf{(1)}] We propose a novel pose-guided feature disentangling transformer for occluded person Re-ID by using pose information to clearly disentangle semantic components (e.g. human body or joint parts) and selectively match non-occluded parts correspondingly.
    \item [\textbf{(2)}] We design a Pose-guided Push Loss to help focus on human body parts and alleviate the interference of occlusion and noise, which avoids the failure of feature learning.
    \item [\textbf{(3)}] To prove the effectiveness of our method, we perform experiments on occluded, holistic Re-ID datasets. Extensive experimental results demonstrate the proposed method performs favorably against SOTA methods.
\end{itemize}


\section{Related Work}
\subsection{Occluded Person Re-Identification}
Occluded person Re-ID is more challenging compared with holistic Re-ID due to body information incompleteness. Existing methods can be basically divided into three categories, hand-craft splitting based methods, methods using additional clues, and methods based on the transformer. 

Methods based on hand-craft splitting handle the occlusion problem by measuring the similarity relationship of the aligned patches. Sun et al. \cite{sun2018beyond} propose a network named Part-based Convolution Baseline (PCB) which uniformly partition the feature map and learn local features directly. Sun et al. \cite{vpm} propose a region based method VPM which perceives the visible region through self-supervision. Jia et al. \cite{MoS} propose MoS which formulates the occluded person Re-ID as a set matching problem by using Jaccard similarity coefficient between the corresponding partten set.

Some methods leverage external cues to locate the human body part such as segmentation, pose estimation or body parsing. Song et al.\cite{song2018mask} propose a mask-guided contrastive attention model to learn features separately from the body. Miao et al. \cite{miao2019pose} introduce Pose-Guided Feature Alignment (PGFA) that utilizes pose information to mine discriminative parts. Gao et al. \cite{PVPM} propose a Pose-guided Visible Part Matching (PVPM) model to learn discriminative part features with pose-guided attentions. Wang et al. \cite{HOReID} propose HOReID that introduces the high-order relation and human-topology information to learn robust features. 

Recently, methods based on transformer are emerging, and the transformer has two major capalities. First, transformer has been proven to have powerful feature extraction capabilities. He et al. \cite{Transreid} investigate a pure transformer framework named TransReID that combines the camera perspective information and achieves good performance on both person Re-ID and Vehicle Re-ID. Second, transformer has the ability to learn the disentangled features. Li et al. \cite{PAT} is the first one to propose Part Aware Transformer (PAT) for occluded person Re-ID, which could disentangle robust human part discovery.

Different from above methods, our method combines the pose information and transformer architecture to clearly disentangle more discriminative features and effectively alleviate the failure of feature learning caused by the occlusion.

\begin{figure*}[t]
    \centering
    \includegraphics[width=1\textwidth]{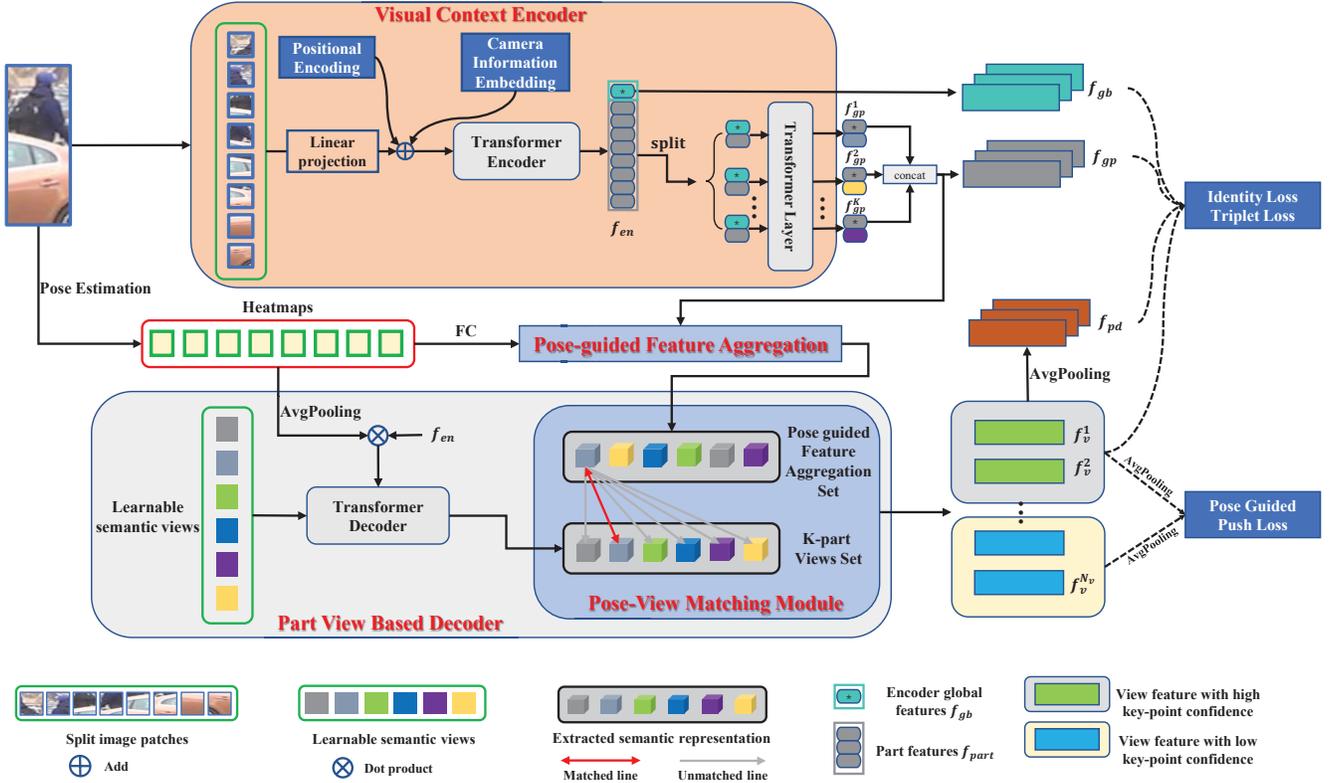}
    \caption{The proposed PFD consists of four parts. The first part is  \textbf{Visual Context Encoder}, which encodes the camera information into embeddings to capture global context information. The second part is \textbf{Pose-guided Feature Aggregation} (PFA) that leverages the matching and distributing mechanism to preliminarily indicate visible body parts. The third part is \textbf{Part View based Decoder} that disentangles the pose-guided feature into discriminative view set under the guidance of the $N_{v}$ learnable semantic views. The fourth part is \textbf{Pose-View Matching module} (PVM), which regards the obtained view set and pose-guided feature set as a set matching problem. In addition, \textbf{Pose-guided Push Loss} is proposed to emphasize the features of visible body parts. For more details, please refer to the proposed method.}
    \label{framework}
\end{figure*}

\subsection{Visual Transformer}
Transformer \cite{vaswani2017attention} has made great achievements in the field of natural language processing. Inspired by the self-attention mechanism, many researchers apply transformers in computer vision. For example, ViT \cite{dosovitskiy2020image} processes images directly as sequences and achieves state-of-the-art performance in image recognition. DETR \cite{DETR} performs cross-attention between object query and feature map to transform detection problem into a one-to-one matching problem, which eliminates the need for hand-crafted components in object detection.

\section{Proposed Method}
In this section, we introduce the proposed Pose-Guided Feature Disentangling (PFD) transformer in detail. An overview of our method is shown in Figure \ref{framework}.

\subsection{Visual Context Transformer Encoder}
We build our encoder based on transformer-based image classification model ($i.e.$ViT\cite{dosovitskiy2020image}). Given a person image $ x \in{\mathbb{R}^{H\times{W\times{C}}}}$, where $H$,$W$,$C$ denote the height, width, and channel dimension respectively. We first split the $x$ into $N$ fixed-sized patches $\{x_{p}^{i}|i=1,2,...,N\}$ by using a sliding window. The step size can be denoted as $S$, the size of each image patch as $P$, and the number of patches $N$ can be described as:
\begin{equation}
N = \lfloor \frac{H+S-P}{S} \rfloor \times \lfloor \frac{W+S-P}{S} \rfloor, 
\end{equation}
where $\lfloor \cdot \rfloor$ is the floor function. When $S$ is equal to patch size $P$, The divided patches are non-overlapping. When $S$ is smaller than $P$, the generated patches are overlapping, which can alleviate the loss of the spatial neighborhood information of the image. The transformer encoder only requires sequence as input, so a trainable linear projection function $f$($\cdot$) is performed on flatten patches to map the patches to $D$ dimensions, and finally obtain the patch embeddings $\textbf{E} \in {\mathbb{R}^{N\times{D}}}$, ($i.e., \textbf{E}_{i} = f(x_{i}),i=1,2,...,N$). A learnable [class] token $x_{class}$ is prepended to the patch embeddings, and the output [class] token serves as encoder global feature representation $f_{gb}$. In order to retain the position information, we apply learnable positional encodings. However, the features are very susceptible to the variation of the camera, so we follow the method in \cite{Transreid} to learn camera perspective information. Then the final input sequence can be described as:
\begin{equation}
    E_{input} = \{x_{class}; \textbf{E}_{i}\} + P_{E} + \lambda_{cm} C_{id},
\end{equation}
where $P_{E}$ is positional embeddings, $C_{id} \in {\mathbb{R}^{(N+1)\times{D}}}$ is camera embeddings, and the $C_{id}$ is same for the same images. $\lambda_{cm}$ is a hyper-parameter to balance the weight of camera embeddings. Next, the input embeddings $E_{input}$ will be processed by $m$ transformer layers. The final output $f_{en} \in \mathbb{R}^{(N+1)\times D}$ of encoder can be divided into two parts (encoder global features and part features): $f_{gb} \in \mathbb{R}^{1\times D}$ and $f_{part} \in \mathbb{R}^{N\times D}$. In order to learn more discriminative features on human parts, the part features $f_{part}$ are split into $K$ groups in order, and the size of each group is $ (N//K) \times D$. Then, each group that concatenates the encoder global feature $f_{gb} \in \mathbb{R}^{1\times D}$ will be fed into a shared transformer layer to learn $K$ group part local feature $f_{gp}=[f_{gp}^{1}, f_{gp}^{2},..., f_{gp}^{K}]$.\\
\textbf{Encoder Supervision Loss.} We adopt cross-entropy loss as identity loss and triplet loss for encoder global features and group features. The encoder loss function can be formulated as:
\begin{equation}
\begin{split}
    \mathcal{L}_{en} &= \mathcal{L}_{id}(\mathcal{P}(f_{gb}))  + \frac{1}{K} \sum\limits_{i=1}^{K} \mathcal{L}_{id}(\mathcal{P}(f_{gp}^{i})) \\ 
    &+ \mathcal{L}_{tri}(f_{gb}) + \frac{1}{K} \sum\limits_{i=1}^{K} \mathcal{L}_{tri}(f_{gp}^{i}), \\
\end{split}
\end{equation}
where $\mathcal{P}(\cdot)$ denotes the probability prediction function.

\subsection{Pose-guided Feature Aggregation}
Occluded person images suffer from less body information, and non-body part information can be ambiguous, which causes performance degradation. Thus, we employ a human pose estimator to extract keypoint information from images.

\textbf{Pose Estimation.} Given a person image $x$, the estimator extract $M$ landmarks from input image. Then the landmarks are utilized to generate heatmaps $\textbf{H}=[h_{1},h_{2},...,h_{M}]$. Each heatmap is downsampled to the size of $(H/4)\times(W/4)$. The maximum response point of each heatmap corresponds to a joint point. We set a threshold $\gamma$ to filter out high confidence landmarks and low confidence landmarks. But unlike \cite{miao2019pose}, we do not set the heatmap whose landmark is smaller than $\gamma$ to 0. Instead, a label $l_{i} \in \{0, 1\}, i=1,...,M$ will be assigned for each heatmap. Formally, heatmap label can be illustrated as:
\begin{equation}
    l_{i} = \left\{
	\begin{aligned}
	0 \quad c_i<\gamma\\
	1 \quad c_i\geq\gamma\\
	\end{aligned} \qquad (i=1,...,M),
	\right
	.
	\label{thresholf gamma}
\end{equation}
where $c_{i}$ denotes the confidence score of i-th landmark.

\textbf{Pose-guided Feature Aggregation.} In order to integrate the the pose information, we set $K = M$, which is exactly equal to the number of keypoints. Then, a fully connected layer is applied to heatmaps $\textbf{H}$ to obtain the heatmaps $\textbf{H}^{'}$, whose dimension is same as the group part local feature $f_{gp}$. Next, the heatmaps $\textbf{H}^{'}$ mutiply $f_{gp}$ element-wisely and obtain the pose-guided feature $\textbf{P}=[P_{1}, P_{2}, ..., P_{M}]$. Although $\textbf{P}$ has explicitly encoded the information of different parts of the human body, we hope to find the part of the information from $f_{gp}$ that contributes the most to a certain body part. Thus, we develop a matching and distributing mechanism, which regards the part local feature and pose-guided feature as a set similarity measurement problem. Finally, we can obtain the pose-guided feature aggregation set $\textbf{S}=\{S_{i}|i=1,2,...,M\}$. For each $P_{i}$, we can find the most similar feature in $f_{gp}$, and then two features are added to form the $S_{i}$. Formally,
\begin{equation}
    k = \mathop{\arg\max}_{j}(\frac{\left\langle P_{i}, f_{gp}^{j} \right\rangle}{||P_i||\ ||f_{gp}^{j}||}),
\end{equation}
\begin{equation}
    S_{i} = P_{i} + f_{gp}^{k},
\end{equation}
where $i=1,2,...,K$, $\left\langle{\cdot,\cdot} \right\rangle$ denotes the inner product, $f_{gp}^k$ denotes the most similar one to $P_i$ in $f_{gp}$.

\subsection{Part View Based Transformer Decoder}
In this section, we define a set of learnable semantic part views to learn the discriminative body parts. The learnable semantic part views can be denoted as $\textbf{Z}=\{Z_{i}|i=1,2,...,N_{v}\}$, and $\textbf{Z}\in\mathbb{R}^{N_{v}\times D}$ will be added to each cross-attention layer as queries. As shown in Figure \ref{framework}, the keys and values are from the combination of pose heatmap $\textbf{H}$ and the output of encoder $f_{en}$. An average pooling layer is applied to heatmap $\textbf{H}$ and then multiply the $f_{en}$, finally output the $f_{de} \in \mathbb{R}^{(N+1)\times D}$. Formally, queries, keys and values can be formulated as:
\begin{equation}
    \textbf{Q}_{i} = Z_{i}\textbf{W}_{q}, \textbf{K}_{j} = f_{de}^{j}\textbf{W}_{k},  \textbf{V}_{j} = f_{de}^{j}\textbf{W}_{v},
\end{equation}
where $i=1,2,...,N_{v}, j=1,2,...,D$, linear projections $\textbf{W}_{q}\in\mathbb{R}^{D\times d_{k}}$, $\textbf{W}_{k}\in\mathbb{R}^{D\times d_{k}}$, and $\textbf{W}_{v}\in\mathbb{R}^{D\times d_{v}}$ are applied to semantic part views and feature $f_{de}$, respectively. Next, we can obtain the $N_{v}$ part views set $v=\{v_{i}|i=1,2,...,N_{v}\}$ by implementing the multi-head attention mechanism and two fully-connected layer, which is the same as \cite{vaswani2017attention}. 

\textbf{Pose-View Matching Module.} In the cross-attention mechanism, the $N_{v}$ part semantic views can learn some discriminative features. However, it is unknown which part or what kind of the information has been learned. Therefore, in order to obtain features related to the human skeleton, we propose a pose-view matching module. Since each feature of the pose-guided feature aggregation set $\textbf{S}$ is related to a certain keypoint information of the human body, we can find the part view $v_{i}$ related to the certain keypoint of the human body by calculating the similarity between the part view $v_{i}$ and $S_{i}$. The matched semantic part view $v_{i}$ and the pose-guided feature aggregation feature $S_{i}$ are added to produce the final view feature set $\textbf{F}_{v}=\{f_{v}^{i}|i=1,2,...,N_{v} \}$. Formally,
\begin{equation}
    k = \mathop{\arg\max}_{j}(\frac{\left\langle v_i, S_j \right\rangle}{||v_i||\ ||S_j||}),
\end{equation}
\begin{equation}
    f_{v}^{i} = v_i + S_k,
\end{equation}
since the confidence score of the landmarks can indicate which part of feature contains human body information, the heatmap label $l_{i}$ can guide us to split the view feature set $\textbf{F}_{v}$ into two parts. Features with heatmap label $l_{i} = 1$ in view set feature form a high-confidence keypoint view feature set $\textbf{F}_{h}= \{f_{h}^{i}|i=1,2,...,L\}$, and the rest form the low-confidence keypoint view feature set $\textbf{F}_{l} = \{f_{l}^{i}|i=1,2,...,N_{v}-L\}$, where $L$ denotes the number of features in $\textbf{F}_{v}$ whose heatmap label is equal to 1.

\textbf{Decoder Supervision Loss.} In order to focus on more non-occluded body features, we propose a Pose-guided Push Loss:
\begin{equation}
    f_{ph} = AvgPooling(\textbf{F}_{h}),
\end{equation}

\begin{equation}
    f_{pl} = AvgPooling(\textbf{F}_{l}),
\end{equation}
\begin{equation}
    \mathcal{L}_{p} = \frac{1}{B} \sum\limits_{i}^{B} \frac{\left\langle f_{ph}^{i}, f_{pl}^{i} \right\rangle}{||f_{ph}^{i}||\ ||f_{pl}^{i}||},
\end{equation}
where $B$ denotes the training batch size. The motivation of this loss is obvious. Human body parts and non-human body parts should not have strong similarities. If $\textbf{F}_{h}$ and $\textbf{F}_{l}$ are similar, then $\mathcal{L}_{p}$ will be large and the learnable sematic part views will adjust themselves adaptively. In order to guide the decoder view feature representation learning, an average-pooling layer is applied to high-confidence keypoints view feature set $\textbf{F}_{h}$ to obtain the pose-guided decoder global feature $f_{ph}$, then identity loss and triplet loss are employed to guide pose-guided decoder global feature $f_{ph}$ and high-confidence feature $f_{h}$ learning as in Eq.\ref{eq13}.
\begin{equation}\label{eq13}
\begin{split}
    \mathcal{L}_{de}  &= \mathcal{L}_{id}(\mathcal{P}(f_{ph}))  + \frac{1}{L} \sum\limits_{i=1}^{L} \mathcal{L}_{id}(\mathcal{P}(f_{h}^{i})) \\ 
    &+ \mathcal{L}_{tri}(f_{ph}) + \frac{1}{L} \sum\limits_{i=1}^{L} \mathcal{L}_{tri}(f_{h}^{i}).
\end{split}
\end{equation}

\subsection{Training and Inference}
In the training stage, the pose estimation uses a pre-trained model, and the rest of components (such as encoder, decoder and so on) are trained together with the overall objective loss, which is formulated as Eq.\ref{overall loss}.
\begin{equation}\label{overall loss}
    \mathcal{L} = \lambda_{en}\mathcal{L}_{en} + \lambda_{de}\mathcal{L}_{de} + \mathcal{L}_{p},
\end{equation}
where $\lambda_{en}$ and $\lambda_{de}$ are the scale factor of encoder loss and decoder loss respectively, and both are set to 0.5.

In the test stage, we concatenate the encoder global feature $f_{gb}$, pose-guided decoder global feature $f_{ph}$, group local part feature $f_{gp}$ and high-confidence keypoint view feature $\textbf{F}_{h}$ as representation $\mathcal{F}$, ingnoring the low-confidence keypoint view feature $\textbf{F}_{l}$. However, high-confidence keypoint view feature $\textbf{F}_{h}$ has variable length, and the network is difficult to implement. Thus, we fix the length of it to $N_{v}$ by padding zeros.
\begin{equation}
    \mathcal{F} = \left[ f_{gb}, f_{ph}, f_{gp}^{1}, ..., f_{gp}^{K}, f_{h}^{1},..., f_{h}^{L} \right].
\end{equation}

\section{Experiments}
\subsection{Datasets and Evaluation Metrics}
 To illustrate the effectiveness of our method, We evaluate our method on five Re-ID datasets for two tasks including occluded person Re-ID and holistic person Re-ID. 
 
 \textbf{Occluded-Duke} \cite{miao2019pose} consists of 15,618 training images, 2,210 occluded query images and 17,661 gallery images. It is a subdataset of DukeMTMC-reID \cite{zheng2017unlabeled}, whichtians occluded images and remove some overlapping images. 
 
 \textbf{Occluded-REID} \cite{zhuo2018occluded} is captured by the mobile phone, which consist of 2,000 images of 200 occluded persons.Each identity has five full-body person images and five occluded person images with different types of severe occlusions. 
 
 \textbf{Market-1501} \cite{zheng2015scalable} contains 1,501 identities observed from 6 camera viewpoints, 12,936 training images of 751 identities, 19,732 gallery images, and 2,228 queries. 
 
 \textbf{DukeMTMC-reID} \cite{zheng2017unlabeled} contains 36,411 images of 1,404 identities captured from 8 camera viewpoints. It contains 16,522 training images, 17,661 gallery images and 2,228 queries. 
 
 \textbf{MSMT17} \cite{wei2018person} contains 125,441 images of 4101 identities captured from 15 camera viewpoints. It contains 32,621 training images. During inference, 82,161 images are randomly selected as gallery and other 11,659 images are considered as query.

\textbf{Evaluation Metrics.} We adopt Cumulative Matching Characteristic (CMC) curves and Mean average precision (mAP) to evaluate the quality of different Re-ID models.

\subsection{Implementation Details}
Both training and testing images are resized to 256 $\times$ 128. The training images are augmented with random horizontal flipping, padding, random cropping and random erasing \cite{zhong2020random}. The initial weights of encoder are pre-trained on ImageNet-21K and then finetuned on ImageNet-1K. In this paper, the number of the split group $K$ and the number of the estimated human landmarks are both set to 17. The number of decoder layer is set to 2 on Occluded-Duke and 6 on the other datasets. The hidden dimension $D$ is set to 768. The transformer decoder is same with \cite{vaswani2017attention}. The batch size is set to 64 with 4 images per ID. The learing rate is initialized at 0.008 with cosine learning rate decay. To detect landmarks from images, we adopt HRNet \cite{sun2019deep} pre-trained on the COCO dataset. The threshold $\gamma$ is set to 0.2.

\subsection{Comparison with the State-of-the-Art}
We compare our method with the state-of-the-art methods on five benchmarks including occluded person ReID and holistic person ReID.

\begin{table}[t]
\small
    \centering
    \begin{tabular}{l|cc|cc}
    \hline
    \multicolumn{1}{l|}{\multirow{2}{*}{Methods}}&\multicolumn{2}{c|}{Occluded-Duke}&\multicolumn{2}{c}{Occluded-REID}\\
    \cline{2-5}
    &  Rank-1  &  mAP   &  Rank-1  &  mAP \\
    \hline
    \hline
    Part-Aligned (ICCV $17$)      &28.8       & 44.6      & -         & -         \\
    PCB (ECCV $18$)               &42.6       & 33.7      & 41.3      & 38.9      \\
    \hline
    Part Bilinear (ECCV $18$)     &36.9       & -         & -         & -         \\
    FD-GAN (NIPS $18$)            &40.8       & -         & -         & -         \\
    Ad-Occluded (CVPR $18$)       & 44.5      & 32.2      & -         & -         \\
    FPR (ICCV $19$)               & -         & -         & 78.3      & 68.0      \\
    PGFA (ICCV $19$)              & 51.4      & 37.3      & -         & -         \\
    PVPM (CVPR $20$)              & 47.0      & 37.7      & 66.8      & 59.5      \\
    GASM (ECCV $20$)              & -         & -         & 74.5      & 65.6      \\
    ISP (ECCV $20$)               &62.8       & 52.3      & -         & -         \\
    HOReID (CVPR $20$)            &55.1       & 43.8      & 80.3      & 70.2      \\
    MoS (AAAI $21$)               &61.0       & 49.2      & -         & -         \\
    \hline
    TransReID (ICCV $21$)           &64.2       & 55.7      & -         & -         \\
    PAT (CVPR $21$)               &64.5       & 53.6      & \textbf{81.6}      & 72.1      \\
    \hline
    \hline
    \textbf{PFD} $(Ours)$           & \textbf{67.7}     & \textbf{60.1}     & 79.8     & \textbf{81.3}  \\
    \hline
    TransReID$^{*}$ (ICCV $21$) &66.4   & 59.2   & -    & -   \\    
    $\textbf{PFD}^{*}$ $(Ours)$     & \textbf{69.5}     & \textbf{61.8}     & 81.5     & \textbf{83.0}     \\
    \hline
    \end{tabular}
    \caption{Performance comparison with state-of-the-art methods on Occluded-Duke, Occluded-REID. "*" means the encoder is with a small step sliding-window setting.}
    \label{tab}
\end{table}

\begin{table}[t]
\small
    \centering
    \begin{tabular}{l|cc|cc}
    \hline
    \multicolumn{1}{l|}{\multirow{2}{*}{Methods}}&\multicolumn{2}{c|}{Market-1501}&\multicolumn{2}{c}{DukeMTMC}\\
    \cline{2-5}
    &  Rank-1  &  mAP   &  Rank-1  &  mAP \\
    \hline
    \hline
    PCB (ECCV $18$)  &92.3   & 77.4   & 81.8    & 66.1   \\
    DSR (CVPR $18$) & 83.6  &64.3     & -       & -     \\
    BOT (CVPRW $19$)  &94.1   & 85.7   & 86.4    & 76.4   \\
    VPM (CVPR $19$)  &93.0   & 80.8   & 83.6    & 72.6   \\

    \hline
    MVPM (ICCV $19$)  &91.4   & 80.5   & 83.4    & 70.0   \\
    SFT (ICCV $19$)  &93.4   & 82.7   & 86.9    & 73.2   \\
    CAMA (CVPR $19$)  &94.7   & 84.5   & 85.8    & 72.9   \\    
    IANet (CVPR $19$)  &94.4   & 83.1   & 87.1    & 73.4   \\
    Circle (CVPR $20$)  &94.2   & 84.9   & -    & -   \\
    \hline
    SPReID (CVPR $18$)  &92.5   & 81.3   & 84.4    & 70.1   \\    
    P$^2$Net (ICCV $19$) & 95.2   & 85.6   & 86.5    & 73.1\\    
    PGFA (CVPR $19$)  &91.2   & 76.8   & 82.6    & 65.5   \\
    AANet (CVPR $19$)  &93.9   & 82.5   & 86.4    & 72.6   \\
    HOReID (CVPR $20$)  &94.2   & 84.9   & 86.9    & 75.6   \\
    \hline
    TransReID (ICCV $21$) &95.0   & 88.2   & 89.6    & 80.6   \\
    PAT (CVPR $21$)     & 95.4    & 88.0   & 88.8    & 78.2   \\
    \hline
    \hline
    \textbf{PFD} ($Ours$)  & \textbf{95.5}   & \textbf{89.6}   & \textbf{90.6}    & \textbf{82.2}    \\
    \hline
    TransReID$^{*}$ (ICCV $21$) &95.2   & 88.9   & 90.7    & 82.0   \\    
    \textbf{PFD}$^{*}$ ($Ours$)  &\textbf{95.5}   & \textbf{89.7}   & \textbf{91.2}     & \textbf{83.2}    \\
    \hline
    
    \end{tabular}
    \caption{Performance comparison with state-of-the-art models on Market-1501 and DukeMTMC-reID datasets.}
    \label{holistic result}
\end{table}

\textbf{Results on Occluded-Duke and Occluded-REID.}
Table \ref{tab} shows the results on two occluded datasets. As table shows, three kinds of methods are compared: (\textbf{1}) hand-crafted splitting based methods including Part-Aligned \cite{zhao2017deeply} and PCB \cite{sun2018beyond}. (\textbf{2}) occluded ReID methods including Part Bilinear \cite{PartBilinear}, PD-GAN \cite{fd-gan}, Ad-Occluded \cite{Ad-occ}, FPR \cite{FPR}, PGFA \cite{miao2019pose}, PVPM \cite{PVPM}, GASM \cite{GASM}, HOReID \cite{HOReID}, ISP \cite{ISP} and MoS \cite{MoS}. (\textbf{3}) Transformer based occluded ReID methods including PAT \cite{PAT} and TransReID \cite{Transreid}. From the table, we can observe that our proposed method PFD achieves 67.7\%/79.8\% Rank-1 accuracy and 60.1\%/81.3\% mAP on Occluded-Duke and Occluded-REID datasets, respectively, and outperforms all kinds of methods in Occluded-Duke. Futher PFD$^{*}$ achieves higher Rank-1 and mAP with a small step sliding-window setting. Compared with PGFA, PVPM and HOReID, which are SOTA methods with keypoints information, our method surpasses them by at least +12.6\% Rank-1 accuracy and +16.3\% mAP on Occluded-Duke dataset. Compared to the competing transformer based methods PAT, our method surpasses it by at least +3.2\% Rank-1 accuracy and +6.5\% mAP on Occluded-Duke and +9.2\% mAP on Occluded-REID.

The reasons for the superior performance of PFD can be attributed to the following points. First, compared with CNN, the transformer has better feature representation ability and can pay better attention to discriminative features. Second, the disentangled features obtained from our method can indicate the body part information in cluttered scenes, leading to clear semantic guidance when matching, which is more effective than spatial alignment. Third, the proposed pose-guided push loss efficiently weakens the interference of occlusions and background clutters.

\textbf{Results on Holistic ReID datasets.}
To verify the effectiveness of our model on the holistic ReID task, we conduct experiments on three holistic ReID datasets including Market-1501, DukeMTMC-reID and MSMT17. Table \ref{holistic result} shows the results on Market-1501 and DukeMTMC-reID datasets. There are four types of methods in the comparison: (\textbf{1}) part feature based methods including PCB \cite{sun2018beyond}, DSR \cite{DSR}, BOT \cite{BOT} and VPM \cite{vpm}. (\textbf{2}) global feature based methods including MVPM \cite{MVPM}, SFT \cite{sft}, CAMA \cite{CAMA}, IANet \cite{IAnet} and Circle \cite{circle}. (\textbf{3}) extra cue based methods including SPReID \cite{spreid}, P$^{2}$Net \cite{p2net}, PGFA \cite{miao2019pose}, AANet \cite{AAnet} and HOReID \cite{HOReID}. (\textbf{4}) transformer based methods including TransReID \cite{Transreid} and PAT \cite{PAT}. From the table, we can observe that our proposed method achieve competitive results. Specifically, our method achieves SOTA performance (95.5\%/90.6\% Rank-1 accuracy and 89.5\%/82.2\% mAP, respectively) on Market-1501 and DukeMTMC-reID datasets. Compared with transformer based method PAT, our method surpasses it by +1.6\% mAP on Market-1501 and +1.8\%/+4\% Rank-1 accruacy/mAP on DukeMTMC. We also conduct experiments on the proposed method on the MSMT17 dataset. Several methods are compared, including MVPM \cite{MVPM}, SFT \cite{sft}, OSNet \cite{OSNet}, IANet \cite{IAnet}, DG-Net \cite{DG-net}, CBN \cite{CBN}, Cirecle \cite{circle}, RGA-SC \cite{RGA-SC}, and SAN \cite{SAN}. From the table \ref{holistic result msmt17} we can see that proposed PFD achieves competitive performance. Specifically, our method achieves 82.7\% Rank-1 accuracy and 65.1\% mAP on MSMT17.
It can be seen that although our method is not designed for holistic reid tasks, it can still achieve competitive results, which reflects the robustness of our proposed method.

\begin{table}[t]
\small
    \centering
    \begin{tabular}{l|cc}
    \hline
    Methods &  Rank-1  &  mAP  \\
    \hline
    \hline
    MVPM (ICCV $19$)    &71.3   & 46.3      \\
    SFT (ICCV $19$)     &73.6   & 47.6      \\
    OSNet (ICCV $19$)   &78.7   & 52.9      \\
    IANet (CVPR $19$)   &75.5   & 46.8      \\
    DG-Net (CVPR $19$)  &77.2   & 52.3      \\
    CBN (ECCV $20$)     &72.8   & 42.9      \\
    Circle (CVPR $20$)  &76.3   & -         \\
    RGA-SC (CVPR $20$)  &80.3   & 57.5      \\
    SAN (AAAI $20$)     &79.2   & 55.7      \\
    \hline
    \hline
    \textbf{PFD} ($Ours$)  &\textbf{82.7}   & \textbf{65.1}    \\
    \textbf{PFD}$^{*}$ ($Ours$)  &\textbf{83.8}   & \textbf{64.4}   \\
    \hline
    
    \end{tabular}
    \caption{Performance comparison with state-of-the-art models on MSMT17.}
    \label{holistic result msmt17}
\end{table}

\begin{table}[t]
\small
    \centering
    \begin{tabular}{c|ccc|cccc}
    \hline
    Index &  PFA &  PVM & $\mathcal{L}_{p}$ &  R-1 &  R-5 &  R-10 &  mAP  \\
    \hline
    \hline
    1   &            &           &            &58.2   & 74.5   & 80.1    & 48.3     \\
    2   &\checkmark  &           &            &63.7   & 77.8   & 82.3    & 56.2     \\
    3   &            &\checkmark &            &62.4   & 76.7   & 81.0    & 54.6     \\
    4   &            &\checkmark &\checkmark  &64.3   & 77.6   & 82.1    & 56.7      \\
    5   &\checkmark  &\checkmark &            &67.0   & 80.0   & 84.4    & 59.5      \\
    6   &\checkmark  &\checkmark &\checkmark  &\textbf{67.7}   & \textbf{80.1}   & \textbf{85.0}   &\textbf{60.1}      \\
    \hline
    \hline
    \end{tabular}
    \caption{Ablation study over Occluded-Duke.}
    \label{ablation study}
\end{table}

\begin{table}[t]
\small
    \centering
    \begin{tabular}{c|cccc}
    \hline
    $N_{v}$ & R-1 &  R-5 &  R-10 &  mAP  \\
    \hline
    \hline
    1   &65.5   & 79.1   & 84.0    & 57.1     \\
    5   &66.7   & 79.9   & 83.7    & 58.4     \\
    10  &66.9   & 79.5   & 83.9    & 58.9     \\
    15  &67.4   & 80.0   & 84.0    & 59.1     \\
    17  &\textbf{67.7}   & \textbf{80.1}   & \textbf{85.0}    & \textbf{60.1}      \\
    20  &66.9   & 79.4   & 84.3    & 59.0      \\
    \hline
    \hline
    \end{tabular}
    \caption{Parameter analysis for the number of semantic views $N_{v}$.}
    \label{semantic views}
\end{table}

\begin{figure}[htbp]
\centering
\begin{subfigure}[]{
\centering
\includegraphics[width=0.43\textwidth]{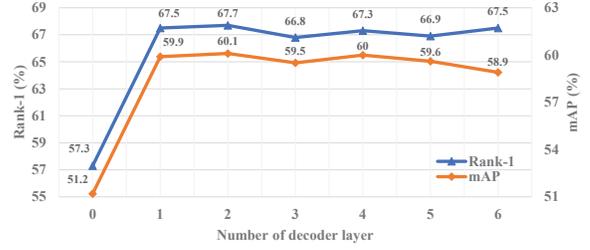}
}
\end{subfigure}

\begin{subfigure}[]{
\centering
\includegraphics[width=0.43\textwidth]{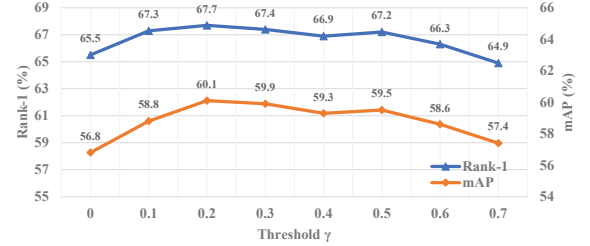}
}
\end{subfigure}
\centering
\caption{Parameter analysis for the number of decoder (a) and the threshold $\gamma$ (b).}
\label{number}
\end{figure}

\subsection{Ablation Study}
In this part, we conduct ablation studies on Occluded-Duke dataset to analyze the effectiveness of each component. 

\textbf{Effectiveness of proposed Modules.} Table \ref{ablation study} shows the experimental results. Index-1 denotes that the pure transformer encoder-decoder architecture. We can see that the performance can reach 58.2\% rank-1 accuracy and 48.3\% mAP, which even shows better performance than pose-guided SOTA method HOReID. This is because the self-attention mechanism can focus on more discriminative features than CNN. From index-2, when pose-guided feature aggregation is added, the performance is greatly improved by +5.5\% rank-1 accuracy and +7.9\% mAP. This shows that the introduction of pose information and correct aggregation can bring good performance improvements. From index-3, we can see that our proposed PVM is also effective. And by comparing index-3 and index-5, we discover that combination of PFA and PVM can increase performance by +8.8\% rank-1 accuracy and +11.2\% mAP, which indicates that pose information and correct matching is very important. From index-5 and index-6, we can see that our overall model can achieve optimal performance, which shows the effectiveness of the Pose-guided Push Loss.

\textbf{Analysis of the number of Semantic views.}
The number of semantic views $N_{v}$ determines the granularity of view features.
As shown in Table \ref{semantic views}, the performance of our proposed PFD is robust to $N_{v}$.
With $N_{v}$ increases, the performance keeps improving before $N_{v}$ arrives 17, which is exactly equal to the number of keypoints. So, we conclude that 17 semantic views may be able to capture the corresponding 17 key point features.

\textbf{Analysis of the number of Transformer Layers.}
 We perform quantitative experiments to find the most suitable number of decoder layer. As shown in Figure \ref{number}(a), when the decoder is removed, the performance of the model is greatly reduced. It can be seen that only the features obtained by the encoder are not robust enough, and the learnable semantic view in the decoder can implicitly learn more important features, which enhances the features from the encoder. we observe that when the number of decoder layer is set to 2, the best performance can be achieved. And with the increase of the number of layers, there is almost no improvement in performance. This is because the resolution of the images in the data set is small, and the content is relatively simple.

\textbf{The Impact of the Threshold $\gamma$.}
The threshold $\gamma$ is defined in Eq \ref{thresholf gamma} to indicate the high confidence landmarks, which could help PVM explicitly match visible body parts. We conduct ablation study on threshold $\gamma$ by changing it from 0 to 0.7.  From the Figure \ref{number}(b), when $\gamma$ is set to 0.2, we can get the best performance. When the value of $\gamma$ is too small, PVM may consider all landmarks as human body areas, thereby introducing noise. Conversely, when the $\gamma$ is too large, a certain body area information may be lost. It is worth noting that when gamma is set to 0, a lot of noise is introduced, but our method can still achieve 65.5\% Rank-1 accuracy, which is still SOTA performance on Occluded-Duke. This shows that our method is robust to pose noise, and further indicates why it could achieve good results on the holistic datasets.

\begin{table}[htbp]
\small
    \centering
    \begin{tabular}{c|cccc}
    \hline
    Methods & R-1 &  R-5 &  R-10 &  mAP  \\
    \hline
    HR-Net(CVPR 19)   & 67.7   & 80.1   & 85.0  & 60.1   \\
    AlphaPose(ICCV 17)   &65.9   & 78.9   & 82.6    & 57.8     \\
    OpenPose(CVPR 17)  &64.1   & 77.8   & 81.2    & 55.6     \\
    \hline
    \end{tabular}
    \caption{Performance of PFD with different pose estimation.}
    \label{diff_pose}
\end{table}
\begin{figure}[htbp]
    \centering
    \includegraphics[width=0.43\textwidth]{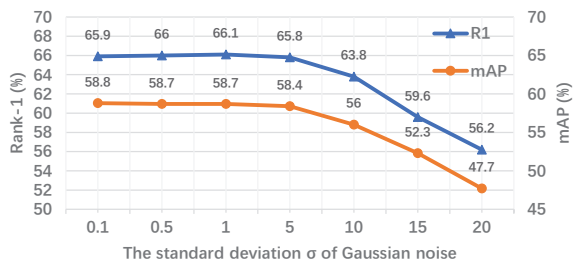}
    \caption{The impact of adding Gaussian noise to the estimated heatmap.}
    \label{noise}
\end{figure}
\textbf{The Impact of Pose Estimation.}
We adopt three different pose estimation algorithms, HRNet \cite{sun2019deep}, AlphaPose \cite{AlphaPose}, and OpenPose \cite{OpenPose} in PFD. From the Table \ref{diff_pose}, the results shows that the PFD still could achieve state-of-the-art performance by using less reliable landmark estimators. Besides, we add Gaussian noise $\mathcal{N}(\mu,\sigma)$ to the estimated heatmap by changing $\sigma$ from 0.1 to 20. From Fig \ref{noise}, we find that the model is robust to pose noise when $\sigma$ is less than 10.

\subsection{Visualization}
\begin{figure}[htbp]
\centering
\subfigure{
\includegraphics[width=0.3in, height=0.6in]{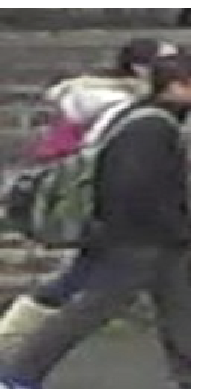}
} 
\centering
\subfigure{
\includegraphics[width=0.3in, height=0.6in]{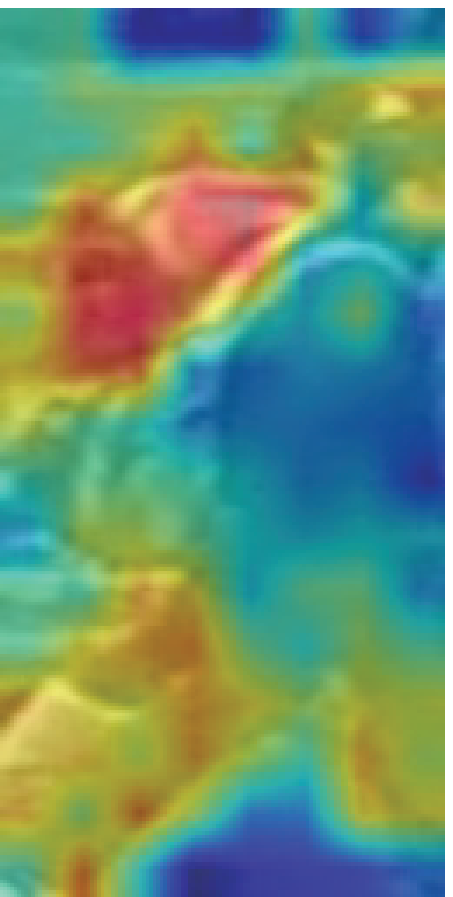}
}%
\centering
\subfigure{
\includegraphics[width=0.3in, height=0.6in]{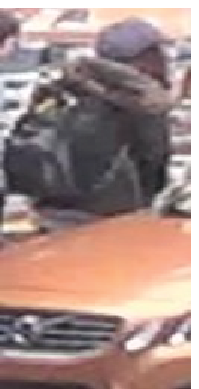}
} 
\centering
\subfigure{
\includegraphics[width=0.3in, height=0.6in]{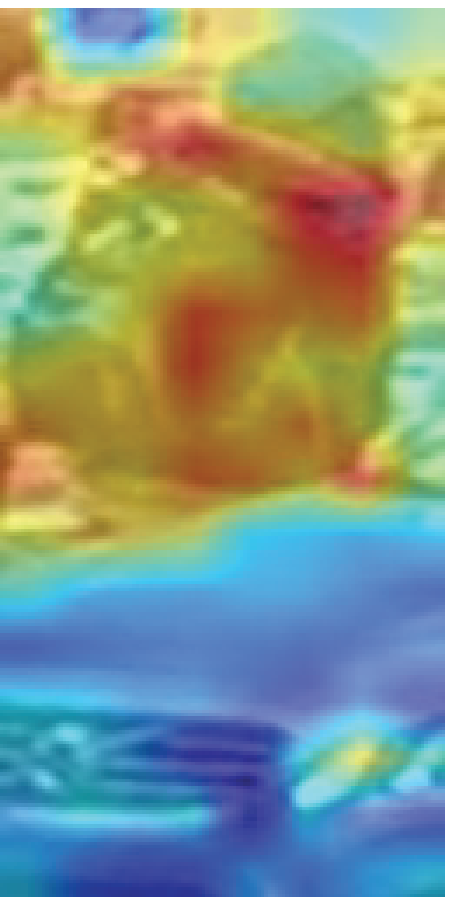}
}%
\centering
\subfigure{
\includegraphics[width=0.3in, height=0.6in]{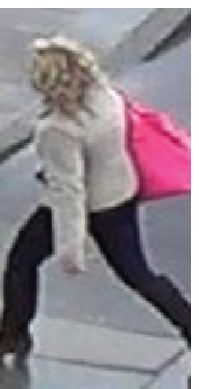}
} 
\centering
\subfigure{
\includegraphics[width=0.3in, height=0.6in]{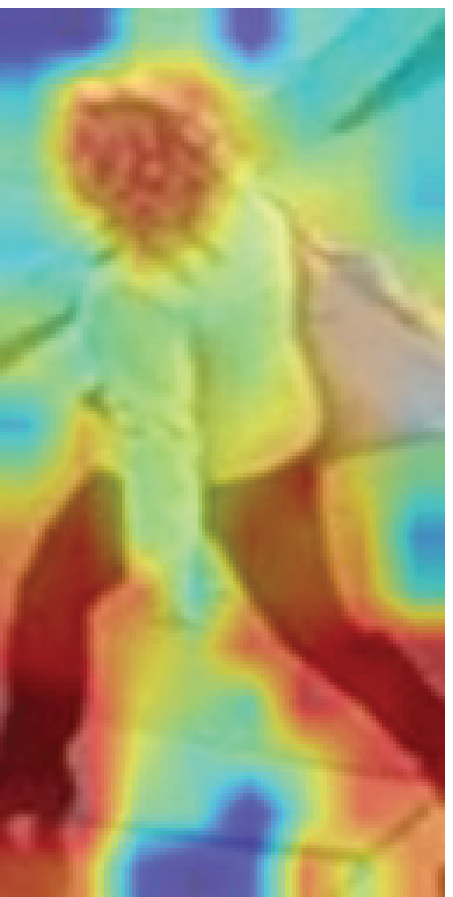}
}%
\centering
\subfigure{
\includegraphics[width=0.3in, height=0.6in]{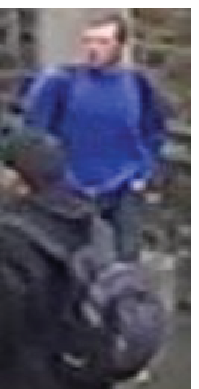}
}%
\centering
\subfigure{
\includegraphics[width=0.3in, height=0.6in]{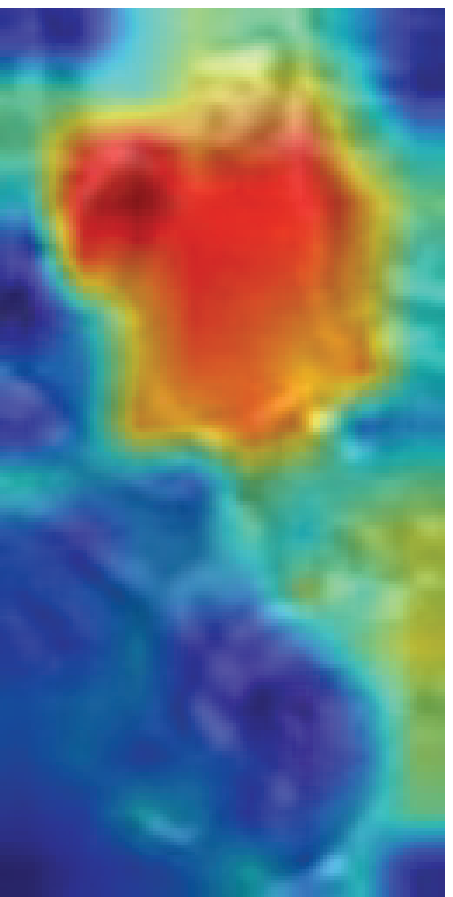}
}%

\centering
\subfigure{
\includegraphics[width=0.3in, height=0.6in]{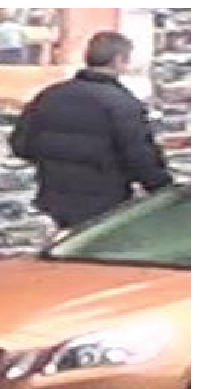}
}
\centering
\subfigure{
\includegraphics[width=0.3in, height=0.6in]{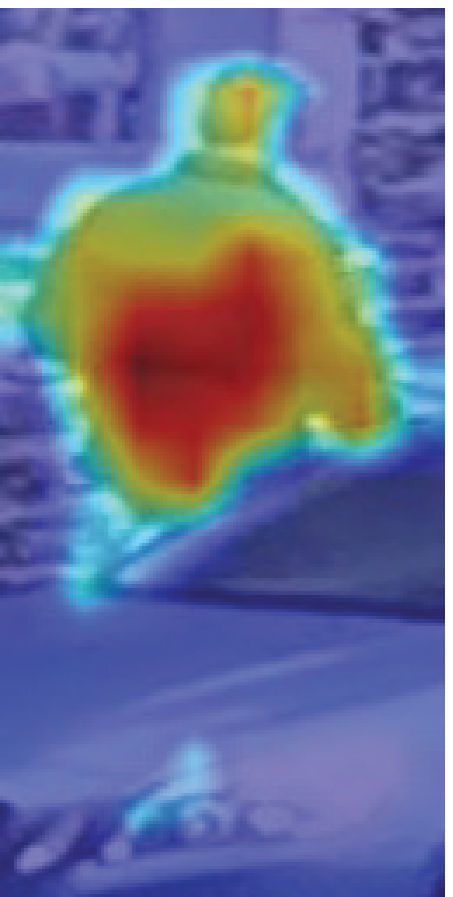}
}%
\centering
\subfigure{
\includegraphics[width=0.3in, height=0.6in]{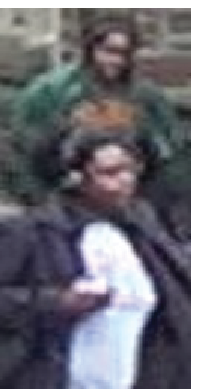}
}
\centering
\subfigure{
\includegraphics[width=0.3in, height=0.6in]{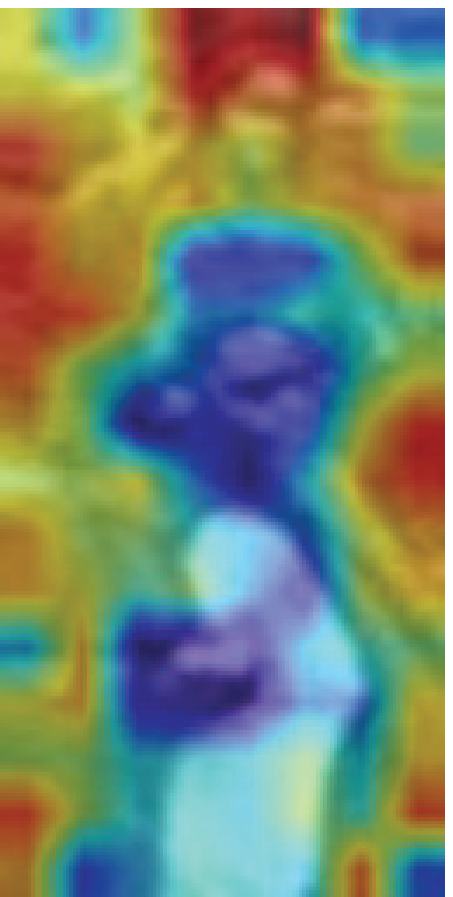}
}%
\centering
\subfigure{
\includegraphics[width=0.3in, height=0.6in]{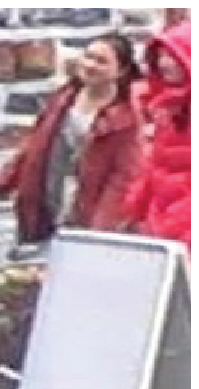}
} 
\centering
\subfigure{
\includegraphics[width=0.3in, height=0.6in]{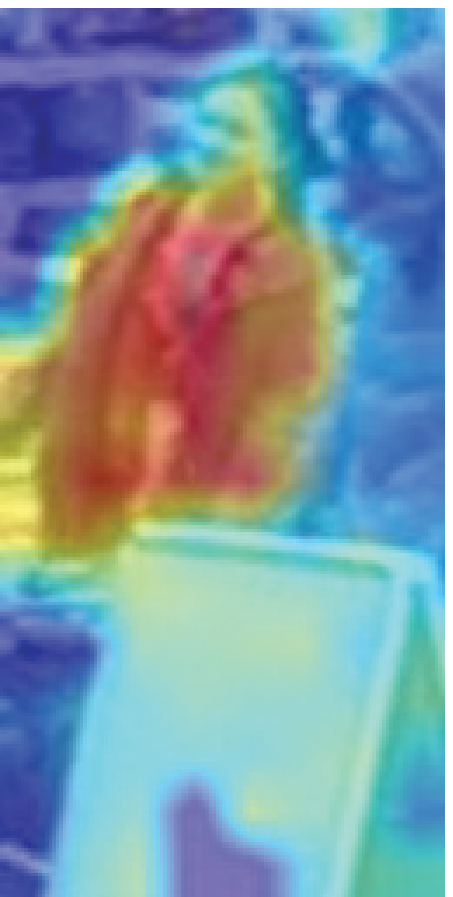}
}%
\centering
\subfigure{
\includegraphics[width=0.3in, height=0.6in]{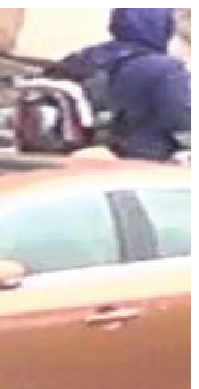}
}%
\centering
\subfigure{
\includegraphics[width=0.3in, height=0.6in]{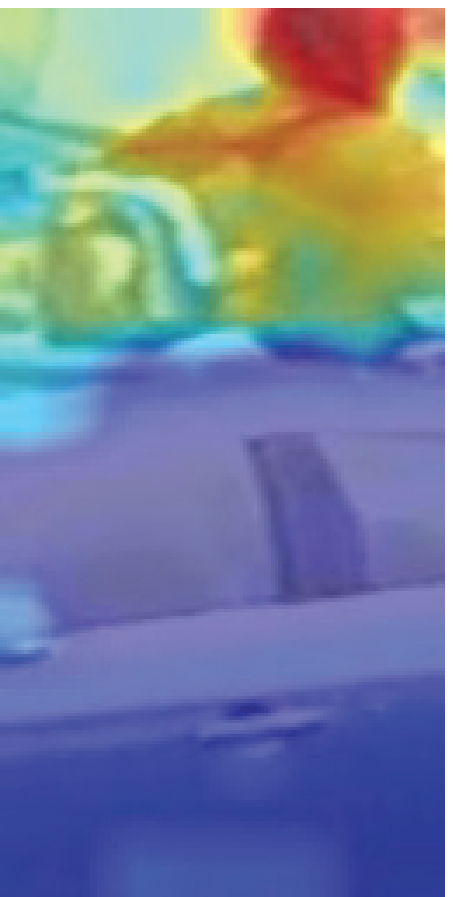}
}%
\centering
\caption{Visualization of decoder attention heatmaps of learned semantic views.}
\label{grad-cam}
\end{figure}

We visualize decoder cross-attention for the different learnable semantic views and fuse them together to form attention heatmap. As Figure \ref{grad-cam} shows, the fused learnable semantic views can almost accurately localize the unobstructed part of the human body, which proves the effectiveness of our proposed method.

\section{Conclusion}
In this paper, we propose a transformer based Pose-guided Feature Disentangling (PFD) method for the occluded Re-ID task that utilizes pose information to clearly disentangle semantic components. PFD contains a transformer based encoder-decoder architecture, two matching modules (PFA and PVM), and a Pose-guided Push Loss. The ViT based encoder extracts the patch features with its strong capability. Then the PFA module preliminarily indicates visible body parts by matching estimated pose heatmaps and patch features. In decoder, we define a set of learnable semantic views to learn the discriminative body parts, and then the PVM module is proposed to enhance the encoder features by matching the most similar features between view set and pose guided feature aggregation set. Besides, PVM cloud automatically separate the occlusion features with the guidance of pose estimation. 
At last, a Pose-guided Push Loss is proposed to better eliminate the interference of occlusion noises by pushing the distance between visible parts and occluded parts in the embedded space. Finally, we conduct experiments on five popular datasets including Occluded-Duke, Occluded-REID, Market-1501, DukeMTMC-reID and MSMT17, and the competitive results demonstrate the effectiveness of the proposed method.


\section{Acknowledgment}
This research was supported by National Key R\&D Program of China (No. 2020AAA0108904), Science and Technology Plan of Shenzhen (No.JCYJ20190808182209321).

\bibliography{ref.bib}

\end{document}